\documentclass[final,3p,times,twocolumn,authoryear]{elsarticle}

\usepackage{nicefrac}
\usepackage{graphicx}
\usepackage{amsmath}
\usepackage{amssymb}
\usepackage{subfig}
\usepackage{hyperref}
\usepackage{makecell}

\DeclareMathOperator*{\argmin}{arg\,min}
\newcommand{\fig}[1]{Fig.~\ref{fig:#1}}

\begin{document}

\begin{frontmatter}



\title{Enhancing Real-World Adversarial Patches through \\3D Modeling of Complex Target Scenes}


\cortext[cor1]{Corresponding author}

\affiliation[inst1]{organization={Department of Software and Information Systems Engineering, Ben-Gurion University of the Negev},
            city={Beer-Sheva},
            postcode={8410501}, 
            country={Israel}}
            
\author[inst1]{Yael Mathov\corref{cor1}}\ead{yaelmath@post.bgu.ac.il}
\author[inst1]{Lior Rokach}\ead{liorrk@bgu.ac.il}
\author[inst1]{Yuval Elovici}\ead{elovici@bgu.ac.il}

\begin{abstract}
Adversarial examples have proven to be a concerning threat to deep learning models, particularly in the image domain.
However, while many studies have examined adversarial examples in the real world, most of them relied on 2D photos of the attack scene.
As a result, the attacks proposed may have limited effectiveness when implemented in realistic environments with 3D objects or varied conditions.
There are few studies on adversarial learning that use 3D objects, and in many cases, other researchers are unable to replicate the real-world evaluation process.
In this study, we present a framework that uses 3D modeling to craft adversarial patches for an existing real-world scene.
Our approach uses a 3D digital approximation of the scene as a simulation of the real world.
With the ability to add and manipulate any element in the digital scene, our framework enables the attacker to improve the adversarial patch's impact in real-world settings. 
We use the framework to create a patch for an everyday scene and evaluate its performance using a novel evaluation process that ensures that our results are reproducible in both the digital space and the real world.
Our evaluation results show that the framework can generate adversarial patches that are robust to different settings in the real world.
\end{abstract}




\end{frontmatter}



\section{Introduction}
The development of deep learning models has contributed to the development of solutions for challenges previously considered unsolvable.
However, a concerning vulnerability in those models was discovered: 
An imperceptible perturbation to a legitimate input sample creates an adversarial example that causes the model to output an incorrect prediction with high confidence \citep{szegedy2013intriguing}.
Although adversarial examples were initially observed in the digital space \citep{goodfellow2014explaining, papernot2016limitations}, they were later demonstrated in the real world \citep{kurakin2016adversarial}; making the threat even greater. 
While several studies have presented various use cases for real-world adversarial perturbations in the image domain \citep{sharif2016accessorize, evtimov2017robust, lee2019physical}, they all utilized a similar methodology: 
First, one or more 2D photos of the target scene are used to craft an adversarial perturbation that can be digitally added to the entire image or applied as a patch to a portion of an image.
Then, the perturbation is recreated in the real world (i.e., printed) and placed in the scene. 
Finally, photos of the scene with the adversarial perturbation are fed to the neural network for evaluation.

\begin{figure}
    \centering
    \includegraphics[width=0.95\linewidth]{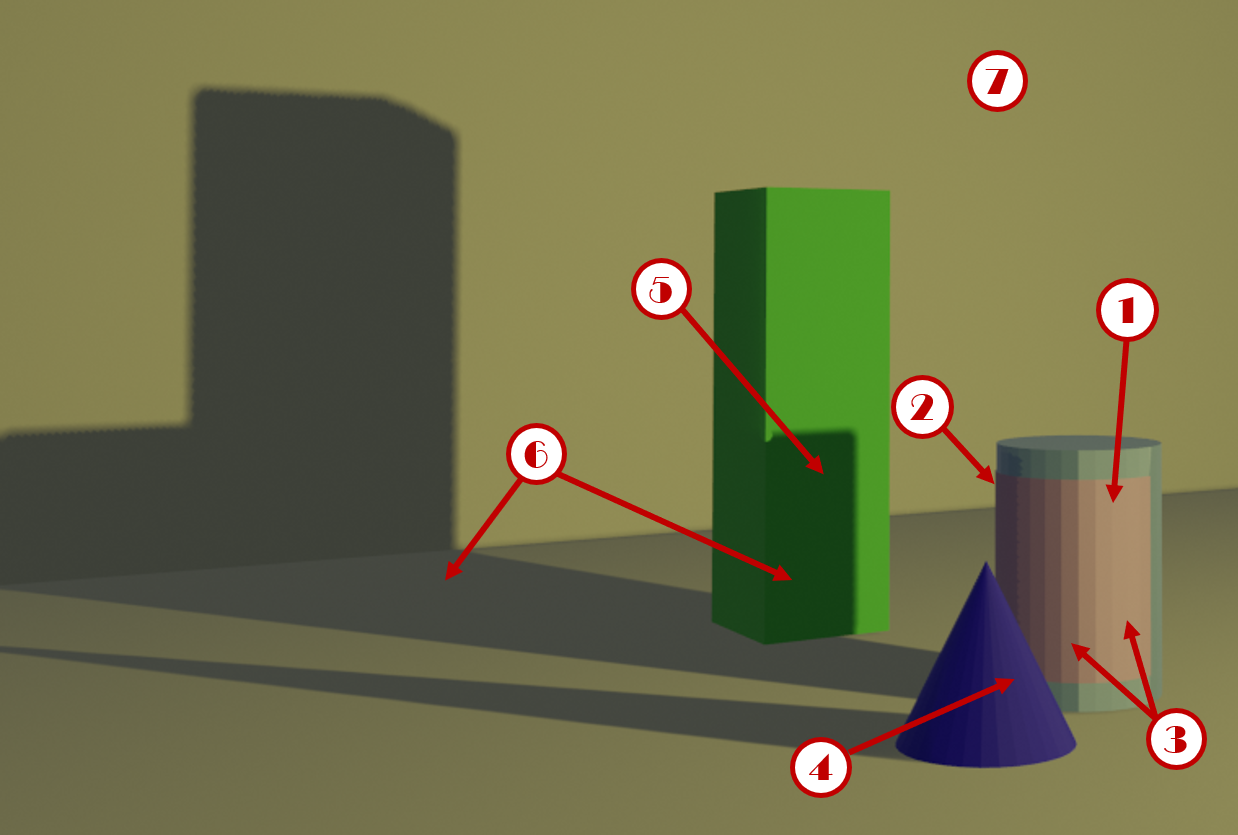}
    \caption{
    Some of the challenges that should be addressed when crafting an adversarial patch for a real-world scene:
    1) The patch is placed on a curved surface. 
    2) The patch is partially hidden by the object it is placed on. 
    3) The patch should have the same lighting as the rest of the scene. 
    4) The patch is hidden by an object that is placed in front of it. 
    5) Objects casting shadows on one another. 
    6) The scene includes an object that affects multiple objects (e.g., an object casts a shadow on more than one object).
    7) The scene may have different environmental conditions (i.e., point yellow light).
    }
    \label{fig:challenges}
\end{figure}

Although that methodology has shown promising results, such 2D image-based methods do not accurately  represent the 3D real world (as shown in \fig{challenges}). 
The main challenge stems from the fact that while both the photos of the scene and the adversarial perturbation are flat, a real-world scene is not.
A flat patch must be placed on a flat surface and must always face the camera; otherwise, parts of the patch will be hidden, and the attack may fail. 
Furthermore, crafting the adversarial perturbation based on photos of the scene limits the attacker's ability to model real-world properties as part of the attack.
Changes to the environmental settings, such as lighting, must be manually added to the real-world scene, and only then can the attacker use them as part of the attack.
In addition, such attacks can only be implemented when the attacker fully controls the target scene, which not a realistic assumption.

Several studies focused on crafting adversarial perturbations for 3D objects. 
A prominent example \citep{athalye2018synthesizing} presents the expectation over transformation (EOT) framework for crafting adversarial perturbations that are robust to random transformations (e.g., rotation, translation); the framework relies on the attacker’s ability to model those transformations as part of the attack process. 
\citet{athalye2018synthesizing} used EOT to perturb the texture of a digital 3D object, which was later printed using a color 3D printer. 
Although successful, the attack targeted a single object, as opposed to the complex scenes that are more common in the real world. 
As \fig{challenges} shows, a realistic scene contains environmental characteristics (e.g., ambient lighting) and different objects that can affect one another (e.g., hide, shadow).
Failing to consider those attributes can impair the attack’s performance. 
Additionally, EOT relies on a specific implementation (limiting the ability to use external modeling and rendering tools) and a 3D color printer, a piece of equipment that is not accessible to many people.
Both issues complicate the attack and make it less feasible.

Another limitation of prior work in this area is the inability to perform the experiment in the real world multiple times and obtain similar results.
In some studies, performing the same real-world experiment more than once can be challenging thus limiting the ability to evaluate the attack.
For instance, it is impossible to replicate and validate an experiment in which a set of photos of the scene was obtained from random positions if the authors do not document the exact positions of the camera.
In other studies, different researchers cannot replicate the results because the target scene or evaluation process cannot be (exactly) recreated by subsequent researchers.
Therefore, in cases like these, the scientific community cannot properly assess methods proposed in prior research on real-world adversarial perturbations.

In this paper, we present a framework that utilizes 3D computer graphic methods to craft adversarial patches that can be added to an existing real-world scene. 
Our approach allows the attacker to target complex realistic scenes, with multiple objects, environmental characteristics, etc.  
The framework uses a digital 3D replica of the target scene to simulate the real world, thus allowing the attacker to assess the limitations of the patch and improve it without risking detection.
Since the attacker controls the digital replica, he/she can add, change, and remove different elements and as a result, can create an adversarial patch that is more robust to real-world transformations.
The framework is designed to allow the integration of external modeling and rendering tools, which gives the attacker the flexibility to implement the attack using the tools he/she prefers.
We implement the framework using open-source tools, and thus the attack is also accessible to an attacker with a limited budget.  

Additionally, we propose an evaluation process that is specifically designed so that the experiment can be performed multiple times and replicated in future studies. 
The two-step evaluation process is used in several experiments to evaluate our attack in both the digital space and the real world.
We also examine whether the use of a digital replica of the scene to create and improve an adversarial patch can simulate the patch's performance in the real world.
To do so, we use the framework to create adversarial patches of multiple target classes and use the evaluation process to compare their performance in the digital space and the real world.
Our results show that using the digital replica to evaluate the adversarial patch can expose useful information about the patch's performance in the real world.
Furthermore, in realistic settings, unexpected changes may occur in the target scene.
Therefore, we examine the patch’s robustness to changes in the scene that were not modeled as part of the scene's replica.
To do so, we evaluate how changing or adding new objects to the scene affects the patch's ability to fool the target neural network.
Finally, by publishing our evaluation setup and code \citep{author_repository}, researchers can reproduce our results and improve upon them. 

The main contributions of this study are:

\begin{itemize}
    \item We present a framework for crafting and improving adversarial patches for an existing real-world scene in the risk-free environment of the digital space.
    \item We demonstrate how the framework can be used to craft a low-budget adversarial patches using free, open-source, and common 3D modeling tools.
    \item We present an evaluation process that enables reproducible experiments in the digital space and the real world.
\end{itemize}

\section{Background}
When discovered by \citet{szegedy2013intriguing}, adversarial perturbations were considered a minor bug, but that changed with the development of advanced adversarial attacks and the growing number of failed attempts to defend against them \citep{goodfellow2014explaining, papernot2016limitations, carlini2017towards}.
Adversarial perturbations' success at fooling neural networks led to an interest in implementing attacks in the real world \citep{kurakin2016adversarial}.
While the first real-world studies performed were unable to reproduce the results obtained in the digital space, these studies set the stage for research on novel forms of adversarial perturbations.
One example is adversarial patches, which are small shapes that, when added to a specific part of an image, could fool object classifiers \citep{evtimov2017robust, brown2017adversarial} or object detectors \citep{lee2019physical, liu2018dpatch}.
Adversarial patches can also be printed and placed in a real-world scene, but their performance is limited.
In all of these cases, the adversarial patches were crafted using a set of 2D photos of the target scene but were used in a 3D space; as a result, the flat patch produced would be unable to address the real-world challenges presented in \fig{challenges}.
Such challenges pushed to the development of methods to improve the robustness of adversarial perturbations in real world settings.

The use of the EOT framework, presented by \citet{athalye2018synthesizing}, to craft perturbations that are robust to specific transformations offered a solution.
This framework builds a set of images by transforming the original sample, using parameters that were randomly sampled from the transformation function's distribution, and use the set to craft a robust adversarial perturbation. 
For example, to craft an adversarial example that is robust to rotation, the training set includes samples of the original image rotated at different angles.
To demonstrate the framework's abilities, EOT was used to perturb the texture of a digital 3D object, which was later printed in the real world using a color 3D printer.
The study focused on perturbing the texture of a single digital object and thus did not consider the challenges of a complex realistic scene (presented in \fig{challenges}).
The EOT framework was demonstrated on a premodeled digital 3D object that was later printed in the real world, however an attacker is more likely to target an existing real-world scene.
Moreover, the implementation of the rendering process and use of a color 3D printer complicate the attack, making it less accessible to inexperienced attackers. 

While some studies examined how the representation of 3D data affects the model's robustness to adversarial perturbations \citep{su2018deeper}, others suggested methods for perturbing the structure (mesh) or texture of 3D objects \citep{Xiang_2019_CVPR, xiao2019meshadv}.
However, the methods focused on manipulating a single 3D object, thus failing to consider the unique characteristics of a complex real-world scene.
\citet{zeng2019adversarial} showed a more realistic approach which perturb different elements of a digital 3D scene to gain more insights about attacks in the real world.
Although using 3D objects to create realistic adversarial examples seems promising, the studies mentioned above used premade digital objects, and did not target an existing complex real-world scene or examine the attack outside of the digital space. 

Our study aims to use the techniques mentioned above to craft adversarial patches that are robust in real-world settings. 
We also improve upon previous research by allowing the attacker to work in a flexible environment that simulates the real world. 
By creating a digital replica of the target scene, we give the attacker control of every element in the scene. 
Then, the attacker can effectively utilize EOT by transforming the digital scene to improve the adversarial patch's robustness to the same transformations in the real world.

\section{Suggested Approach}
\subsection{Assumptions and Threat Model}
We assume an attacker that wants to add a sticker to an existing real-world scene to hide an object from an object detection system that uses a deep learning model. 
By doing so, all photos of the scene will result in a false prediction by the neural network.
We assume the attacker has complete knowledge of the target neural network (e.g., parameters, architecture) but not the other system components (e.g., camera).
Moreover, the attacker can examine the real-world scene to create a 3D digital replica of it, such that the neural network classifies the rendered images of the replica as the original class.
Finally, we assume that the attacker has physical access to the actual scene to add the patch in the real world.
While this study demonstrates an attack under white-box settings, our framework can use black-box attack methods to craft an adversarial patch under more restrictive settings.

\subsection{Framework Overview} 
\begin{figure*}
    \centering
    \includegraphics[width=0.9\linewidth]{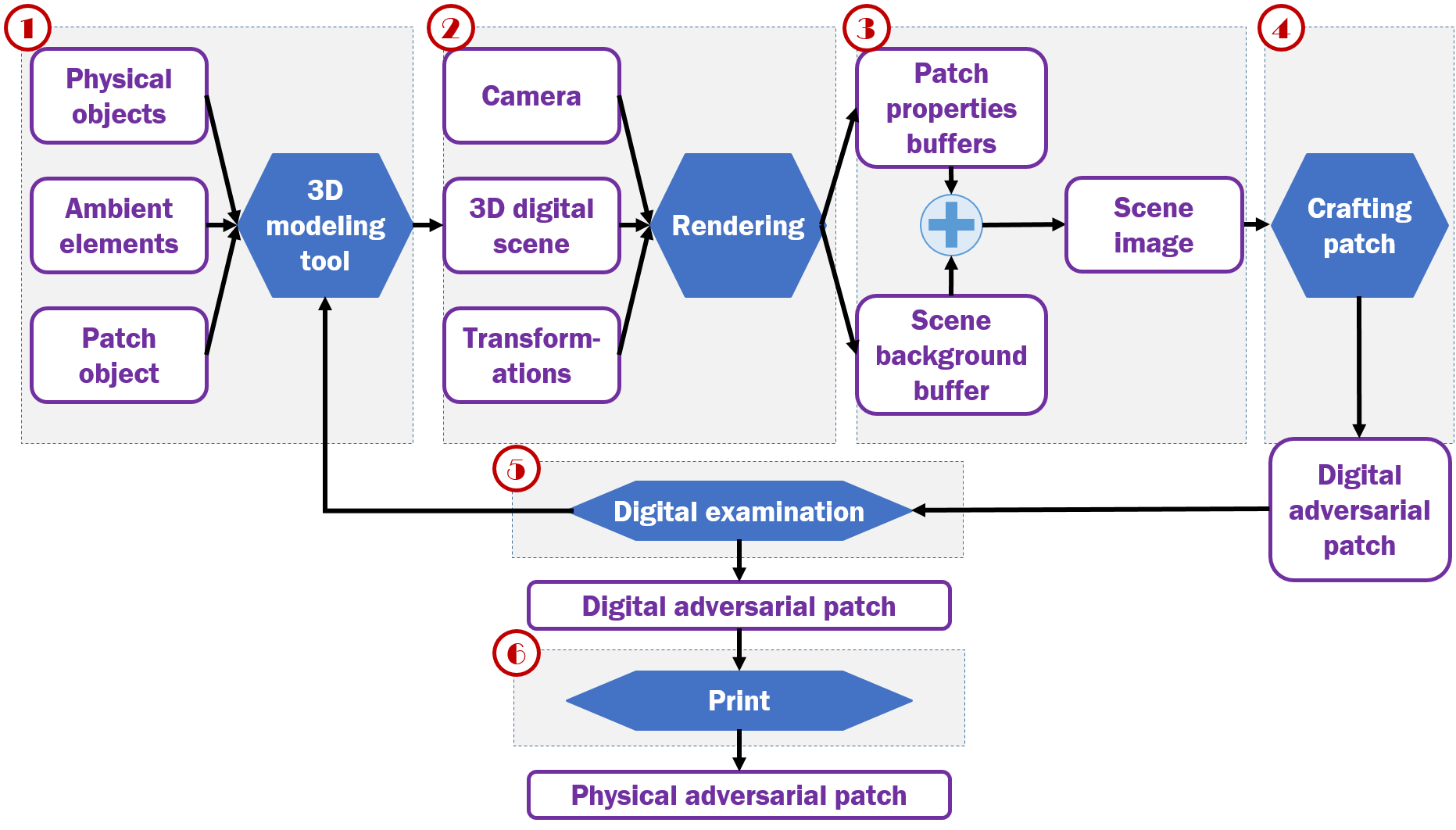}
    \caption{
    The framework’s steps: 
    1) Model a digital replication for the real-world scene. 
    2) Render 2D images with realistic transformations. 
    3) Combine the output of the rendering process into differentiable images.
    4) Craft the adversarial patch.
    5) Examine the adversarial patch in the digital space and improve the attack if needed. 
    6) Apply the patch to the physical scene.
    }
    \label{fig:framework}
\end{figure*}

We suggest the following framework for crafting an adversarial patch for an existing real-world scene. 
\fig{framework} presents the six steps of the framework: model, render, combine, craft, evaluate, and apply.
The first five steps are performed in the digital space, while in the sixth step, the adversarial patch is transferred from the digital space to the real world.  

First, the attacker uses 3D modeling techniques to create a 3D digital replication of the real-world scene and adds an empty digital patch object to the replica; in the following steps the texture of the empty object will be perturbed to craft the adversarial patch.
Second, the digital replica is rendered into 2D images, such that each image is a rendering of the scene under real-world transformations (e.g., rotation, a change in the lighting).
The rendering process results in two outputs: a background image of the scene (without the patch) and the properties of the patch (all pixels that include the patch). 
Third, the attacker uses a differentiable method to combine the background and patch properties into one image to create a set of the scene’s images.
This way, the attacker can use a wide variety of external rendering tools without implementing a differentiable rendering process or approximating the gradients of a non-differentiable rendering. 
Fourth, the images are then used to perturb the patch’s texture, using any method of crafting an adversarial perturbation. 
Fifth, the attacker adds the adversarial patch to the scene’s digital replica, renders images of the scene, and feeds them to the neural network. 
As a result, the attacker can examine the patch’s effect on the scene and improve it if needed. 
Finally, the attacker prints the final patch on a sticker and adds it to the real-world scene.

\subsubsection{Modeling a Digital Replica of the Real-World Scene}
The first step is to create a 3D digital replica that approximates the real-world target scene. 
To build this replica, the attacker can choose any tool (e.g., Blender, Maya) or resources (e.g., use free 3D objects, buy 3D designs, use a 3D scanner). 
Then the attacker adds an empty 3D object to the digital scene to serve as the adversarial patch in the replica. 
In this step, the attacker should consider the following elements: the objects in the scene, the ambient characteristics, and the adversarial patch.
First, the replica is built from digital 3D objects that represent the objects in the real-world scene. 
Then, the scene’s ambient characteristics (e.g., light sources, smoke) are added to the replica to improve the replica's similarity to the real world. 
Finally, the digital patch object is added to the scene according to its expected location in real life.
Decisions regarding the location, shape, and size of the patch should be made based on the attacker’s goals and the camera’s expected location.

The modeling stage is affected by the expertise and resources of the attacker.
Our findings show that a successful attack can be launched by roughly approximating the target scene, however when more realistic replications are modeled, the success rate will likely increase. 
This finding demonstrates a trade-off between the attacker’s effort and the success rate, which can be used to balance the goals and capabilities of the attacker. 

\subsubsection{Rendering 2D Images with Realistic Transformations}
The EOT process uses a collection of images with different transformations (“views”) to craft a single perturbation that fools the target neural network for all views. 
Unlike past studies that could only use EOT with a limited number of transformation functions, our framework's design allows the attacker to transform any of the scene’s properties, including the scene’s objects (one or more), ambient characteristics, camera view, and more.
As presented by \citet{athalye2018synthesizing}, flexibility in choosing the transformations results in a better adversarial perturbation.
Therefore, in this step, the attacker aims to create a collection of views, such that each view represents the digital replica under a different set of transformations.

We define $T = \{ T_1,...,T_k \}$ such that each $T_i$ $(1 \leq i \leq k )$ is a distribution of transformation functions $t_i$ on the digital replica (i.e., the scene).
For example, a transformation function that rotates the scene through an angle $\theta$ around the x-axis can be sampled from $U(30,90)$; hence, $\theta \in [30,90]$.
Similarly, we define $C$ as the distribution for the transformation functions on the digital replica's camera (e.g., the camera's position).
For each set of transformation functions sampled from $T$ and $C$, the rendering process $R$ applies the transformations to the 3D digital replica $S$ with patch texture $P$ and outputs a 2D image of the 3D scene with the $t$ transformations from viewpoint $c$ (a view).
In this step, the attacker samples transformation functions multiple times to build a collection of views $X$:
\begin{equation}
    \nonumber
    X = \mathbb{E}_{t \sim T, c \sim C} [R (S,P,t,c)]
\end{equation}

\subsubsection{Combining the Rendering Output Buffers} 
In the next step (i.e., crafting), the set of views is used to perturb the patch's texture $P$ to create the adversarial patch.
As shown in previous studies, crafting the adversarial perturbation is usually done by solving an optimization problem by calculating or approximating the attack’s loss gradient concerning the patch. 
Because our framework allows the use of external rendering tools, an attacker can use a non-differentiable rendering process to craft $X$.
As a result, the attacker cannot calculate $\frac{\partial X}{\partial P}$, those preventing him/her from using gradient-based methods for crafting adversarial perturbations.
Past studies overcome this issue by implementing the rendering process as a differentiable part of the attack \citep{athalye2018synthesizing, Xiang_2019_CVPR} or by approximating the loss gradient \citep{xiao2019meshadv, zeng2019adversarial}. 
However, these solutions limit the attacker from using external tools and increase the knowledge required to implement an attack.

Therefore, we suggest modifying the output of the rendering process, which is a simple configuration change that can be done in almost any rendering tool.
Then, instead of outputting a single image, the result of the rendering process is multiple buffers that can be split into two types: the background and the patch’s properties. 
The background is an image of the scene without the patch, and the patch’s properties form a set of buffers with information that allows the patch to be added to the background in a realistic manner; combining the two creates an image of the scene.
Hence, the rendering stage results in a set $X=\{(b_i,p_i)\}_{i=1}^n$, where for sample $x_i=(b_i,p_i) \in X$, $b_i$ is the scene background, and $p_i$ are the patch properties. 
Additionally, for a sample $x_i$, let $B(P,b_i,p_i)$ be a differentiable method that combines $P$, $b_i$, and $p_i$ into an image of the rendered scene. 
Given the result of the rendering step $X$, the attacker builds a set of views $\tilde{X}$ that are differentiable by $P$:
\begin{equation}
    \nonumber
    \tilde{X} = \{ B(P,b_i,p_i): \forall (b_i,p_i) \in X \}
\end{equation}

\begin{figure*} 
    \centering
    \null\hfill
    \subfloat[Sampling the texture]{\label{fig:combining-sample}
        \includegraphics[width=0.40\linewidth]{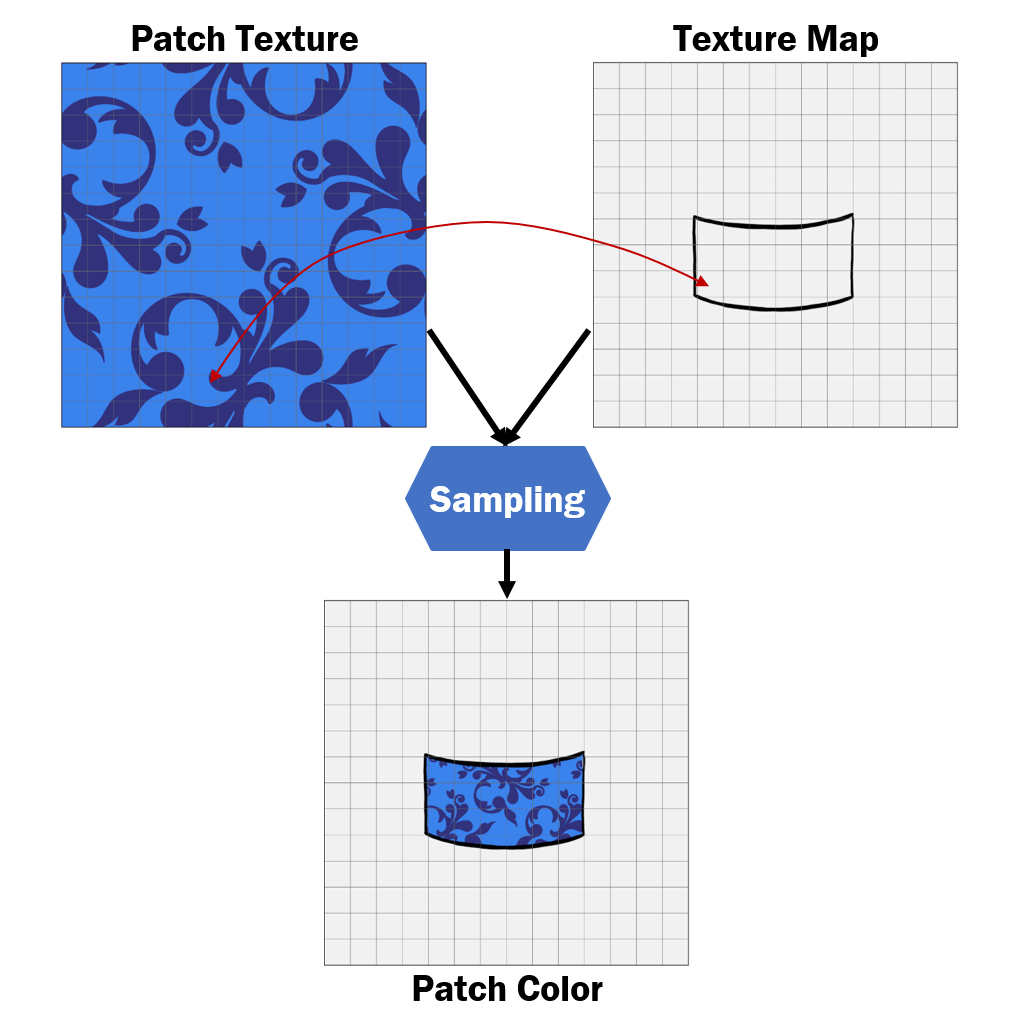}}\hfill
    \subfloat[Combining the buffers]{\label{fig:combining-buffers}  
        \includegraphics[width=0.40\linewidth]{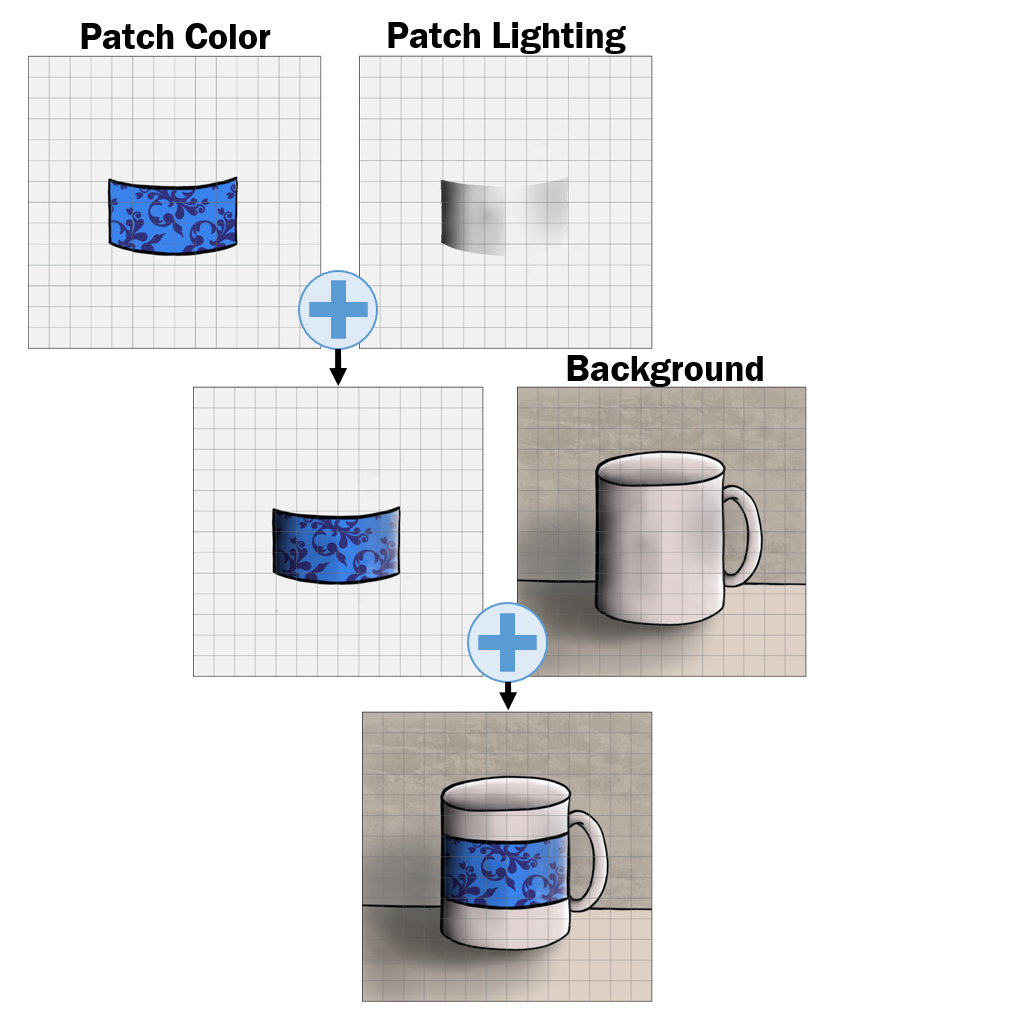}}
    \hfill\null
    
    \caption{
        Creating the images of the scene from the patch properties and background buffers.
        (a) Sampling the texture of the patch to create an image of the patch's colors.
        (b) Combining three buffers to create an image of the scene: the patch's colors, the shading of the patch, and an image of the background without the patch. 
    }
    \label{fig:combining}
\end{figure*}

An example of combining buffers into a scene image is presented in \fig{combining}.
The background buffer $b_i$ is an image of the scene without the patch, and the two patch properties buffers $p_i$ are the patch's texture map and lighting.
Each pixel in the texture map contains the coordinates of a pixel in the patch's texture $P$, thus allowing the framework to build an image of the patch's colors by sampling pixels from $P$ (shown in \fig{combining-sample}).
Then, the patch's color and lighting and the scene background are combined into one complete image of the complete scene (shown in \fig{combining-buffers}).
Since the image is built using simple operators, like sampling $P$, addition, and multiplication, the image $B(P,b_i,p_i)$ is differentiable by $P$. 

\subsubsection{Crafting the Adversarial Patch}
After combining the buffers into a set of images $\tilde{X}$, the attacker can use them to craft the adversarial patch.
Crafting an adversarial perturbation under white-box settings is usually done by feeding a benign input sample to the target learning model, calculating both the attack loss based on the model’s output and the loss gradient with respect to the input sample, and perturbing the original input sample.
The attack loss is used to ensure that the adversarial perturbation meets the attack's requirements, such as fooling the target model to output a specific result, limiting the perturbation size, etc.
In most attacks, the original input sample is gradually changed by an iterative optimization process that adds a small perturbation to the sample in each iteration.
While most attacks rely on a similar methodology, they differ by parameters and configuration (e.g., the attack loss, optimization process, number of iterations).
Given our framework's design, most methods for crafting adversarial perturbations can be used in this step to perturb the texture of the empty patch digital object into an adversarial patch.

Since combining the rendering output buffers results in a set of images $\tilde{X}$ that are differentiable by the patch texture $P$, the attacker can use any gradient-based method to craft an adversarial perturbation to $P$.
Most methods perturb the patch texture $P$ to fool a neural network $M$ by optimizing an objective function with a customized attack loss $\mathcal{L}$, e.g., finding $P$ such that $\argmin_P(\mathcal{L}(M(\tilde{X})))$.
The attacker can construct $\mathcal{L}$ to create a targeted attack, apply constraints, etc. 
Then, since it is easy to calculate $\frac{\partial \tilde{X}}{\partial P}$, the attacker can also calculate $\frac{\partial \mathcal{L}(M(\tilde{X}))}{\partial P}$ to solve the optimization problem with a gradient-based optimizer (e.g., gradient descent). 
As done in previous studies, the objective function and optimization method should be selected according to the attack's goal and target neural network. 
While in this study we demonstrate our framework with a white-box attack, future work can replace the attack method we used in this step with a black-box attack (such as \citep{chen2017zoo, chen2020hopskipjumpattack}) to craft an adversarial patch under more restrictive settings.

\subsubsection{Examining the Patch in the Digital Space}
Next, by adding the patch to the digital replica, the attacker can simulate the adversarial patch’s effect on the neural network’s prediction concerning the real-world scene. 
Examining the adversarial patch in the real world requires both the attacker's presence at the original scene and the performance of actions that could be considered abnormal thereby exposing the attacker to the risk of detection.
Using the digital replica to simulate the real-world scene allows the attacker to identify potential problems and improve the patch in a controlled and safe environment.
Since the attacker controls the digital replica, he/she can simulate events that are challenging to control in real life (e.g., waning daylight, the presence of smoke). 

Moreover, the attacker can use the digital replica to compare different adversarial patches, identify the most effective one, and improve the attack's success in the real world.
Based on the findings in this step, the attacker might choose to change the attack process, requiring modifications to the scene’s replica, the addition of new transformations to the rendering step, or changes to the attack’s optimization function.  
Since the examination process is performed in the digital space, the attacker can improve and evaluate the patch as long as the attacker wishes.

\subsubsection{Applying the Patch to the Physical Scene}
Finally, the attacker creates the adversarial patch in the real world and adds it to the scene. 
For instance, the patch can be printed on a sticker or a piece of paper using a home printer.

\section{Experimental Setup}
\begin{figure*}
    \centering
    \null\hfill
    \subfloat[Real world]{\label{fig:mugs-rw}
        \includegraphics[width=0.40\linewidth]{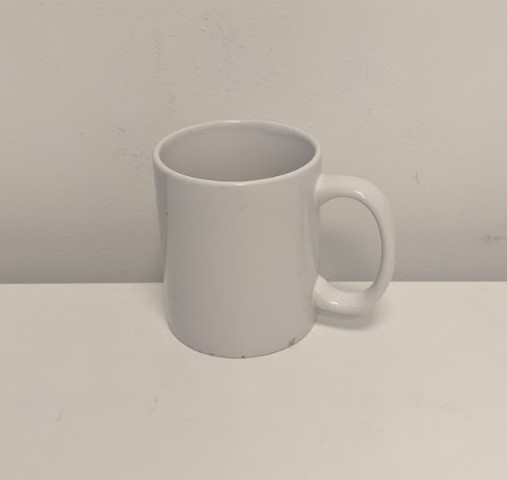}}\hfill
    \subfloat[Digital replica]{\label{fig:mugs-digital}
        \includegraphics[width=0.40\linewidth]{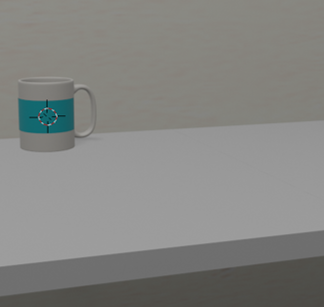}}
    \hfill\null
    
    \caption{
        The target scene: a white mug placed on an office desk; (a) is a photo of the original real-world scene; and (b) is a rendering of our digital replica with the empty adversarial patch (blue strip) that was modeled using Blender. 
    }
    \label{fig:mugs}
\end{figure*}

In this study, we implement the framework using free and open-source software to create an adversarial patch for a typical office scene in which a standard white mug is placed on a desk, as seen in \fig{mugs-rw}. 
A webcam (Microsoft LifeCam VX-700) films the scene, and then we divide the video stream into photos, crop them into $299 \times 299$ pixel color images, and feed them to a state-of-the-art object classifier, InceptionV3 \citep{xia2017inception}. 
This classifier was trained on the ImageNet dataset \citep{deng2009imagenet}, achieving 1-top accuracy of 78\% and 5-top accuracy of 93.9\% for valid input. 
We validated that the scene is classified as the original class (i.e., Coffee Mug) for 100\% of the images rendered from the digital replica without the patch.
Additional information on the experiment setup, code, and parameters is available in \citep{author_repository}.

\subsection{Crafting the Adversarial Patch}
\subsubsection{Modeling}
To model the replica of the target scene, we use Blender \citep{blender}, a free creation software. 
We start by approximating the objects in the real-world scene: 
For the desk and the walls of the room, we use default cube shapes in Blender. 
However, the other objects are more complex, and a professional expertise in 3D modeling is required to create them from scratch.
Therefore, we use a free, premade mug object \citep{coffeemugmodel} and add it to the digital replica.
Afterward, we add a yellow point light source that simulates the light bulb in the original office.
Then, to create the empty placeholder for the adversarial patch, we crop and edit a part of the mug's model \citep{coffeemugmodel} and the result around the mug's 3D object in the digital scene.
For each 3D object, we choose standard configurations for the materials and use textures made from photos of the real-world scene.
We note that the replica is modeled using online tutorials for beginners to produce a simple approximation of the real-world scene.
Therefore, while our digital scene lacks some realistic elements (\fig{mugs-digital}), it can be easily replicated by attackers and researchers with limited experience in 3D modeling.

\subsubsection{Rendering} 
The rendering process is implemented with ModernGL \citep{moderngl}, based on the examples in the library’s repository, and uses the configurations and materials to portray the ambient elements in the rendered images (e.g., shadows).
Due to the attacker's assumed lack of knowledge about the camera, we estimate the camera's configurations (e.g., field of view).
Then, we render a set of views $X$, i.e., images of the scene under different transformations:
For the scene transformations $T$, we choose translation and rotation in the $x, y,$ and $z$ axes, and changes in the light color, and for the camera transformation $C$, we choose changes in the camera's position. 
To determine the ranges that define each distribution of the different transformation functions, we examine both the digital replica and the real-world scene.
Based on our findings, we define the ranges according to possible changes in the real-world scene while ensuring that the mug and patch are visible in all views.
Additional information about the rendering step (including parameters and code) is available in \citep{author_repository}. 

In this study, we explore two methods for sampling the parameters for the transformations: \textit{random sampling} and \textit{systematic sampling}. 
Random sampling is commonly used in the EOT framework, in which a transformation function $t_i$ is randomly sampled from a uniform distribution $T_i \sim U(\alpha_i,\beta_i)$; hence, the parameter that defines $t_i$ is randomly sampled from the range $[\alpha_i,\beta_i]$.
Additionally, we examine a deterministic approach, which we refer to as systematic sampling, where the transformation functions $t_i$ are predefined by systematically sampling the function's parameter in constant even steps across $[\alpha_i,\beta_i]$.
For example, we want to create $l$ scene rotation functions $\{ t_{i,1}, ...,t_{i,l}\} \in T_i$ that are systematically sampled from the range $[\alpha_i,\beta_i]$.
Therefore, we predefine $\{ t_{i,1}, ...,t_{i,l}\}$ such that for each $1 \leq j \leq l$, the rotation function $t_{i,j}$ rotates the scene through an angle of $\theta_j = \alpha_i + \frac{j \cdot (\beta_i - \alpha_i)}{l}$.
After selecting the transformation functions, we build the set of views by rendering an image of the scene using every combination of those parameters for the different transformation distributions.

\subsubsection{Combining}
The rendering step outputs a set of views $X$ that is determined by the sampling method; hence, each view is defined by a set of scene and camera transformations. 
Additionally, for each view, the rendering process outputs four buffers that can be split into two types: one buffer with background image and three buffers with the patch properties.
The patch information buffers include the texture mapping, lighting, and a mask that defines the parts of the patch object visible in the rendered image.
To build a differentiable image from the four buffers, we follow a similar process to the one presented in \fig{combining}: 
We use the texture mapping to sample the colors from $P$, merge the result with the patch's lighting, and finally, use the mask to combine the background image with the patch.
The result is an image of the complete scene with the patch, similar to the one shown in \fig{mugs-digital}.
The building process is implemented using TensorFlow \citep{abadi2016tensorflow} and uses simple operations such as sample, add, and multiply tensors; as a result, the output image is differentiable by $P$.
We use this process for each view in $X$ to build a set of differentiable images $\tilde{X}$.

\subsubsection{Crafting} 
\begin{figure}[t] 
    \centering
    \includegraphics[width=0.85\linewidth]{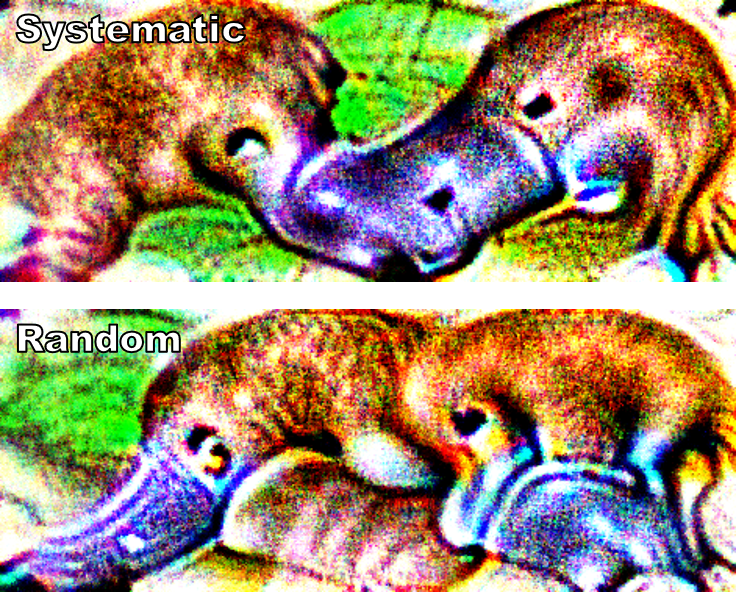}
    \caption{
        The systematic (upper) and random (bottom) adversarial patches for the Platypus target class. 
    }
    \label{fig:patch}
\end{figure}

To craft the adversarial patch, we follow previous studies and define an objective function with a customized attack loss:
\begin{equation}
    \nonumber
    \mathcal{L}(\tilde{X},P) = CE(\tilde{X}, y_{tg}) -\kappa \cdot CE(\tilde{X}, y_{og}) +\lambda \cdot TV(P) 
\end{equation}
where $\kappa$ and $\lambda$ are tuning parameters, $y_{og}$ and $y_{tg}$ are the original and attack target class respectively, $CE$ is the cross-entropy loss, and $TV$ is the total variation.
Then, to perturb $P$, we use the Adam optimizer to solve the following optimization problem:
\begin{equation}
    \nonumber
    \argmin_P \{ \mathcal{L}(\tilde{X},P) \}    
\end{equation}
The attack loss $\mathcal{L}$ uses the cross-entropy loss to cause the scene with the patch to be classified as $y_{tg}$ and not as $y_{og}$, while maintaining the smoothness of the patch by minimizing the total variation distance.
Examples of two adversarial patches are presented in \fig{patch}.


\subsubsection{Examining} 
To examine the patch, we use the digital evaluation process, which is described in Section~\ref{sec:evaluation}.
Based on our findings, we perform several changes in the transformation ranges and updates to the 3D replica to improve it (e.g., creating realistic lighting by modeling the room as a box instead of the three visible walls).
The attack was initially designed for the Armadillo target class; later, we change the seed and initialization parameters and use the attack to craft patches for the Armadillo and nine additional target classes.
We use the same attack to perform a non-biased comparison between the patches for the different target classes.
However, it is more realistic to build a tailored attack for each target class, an approach we will explore in future research.

\subsubsection{Applying} 
We print the patch on an A4 piece of paper using a Xerox WorkCentre 6605, manually crop it, and apply it to the mug using transparent adhesive tape.

\subsection{Evaluation Process}\label{sec:evaluation}
The ability to perform the same experiment multiple times under the same conditions and obtain similar results is essential for our study’s integrity and to enable future research to reproduce our work. 
Therefore, we present a replicable two-step evaluation process, the first step of which takes place in the digital space; the second step takes place in the real world. 
It is important to note that the evaluation process was designed for research purposes and is not part of our framework.

\subsubsection{Digital Space}
In the first step, the digital replica is used to simulate the evaluation in the real world, similarly to our framework's examination step (step 5 in \fig{framework}).
Here, we add the adversarial patch to the scene, render a set of images under the expected real-world settings and transformations (e.g., camera's position), send the images to the neural network, and analyze the predictions.

\subsubsection{Real-World} 
\begin{figure*}
    \centering
    \includegraphics[width=0.9\linewidth]{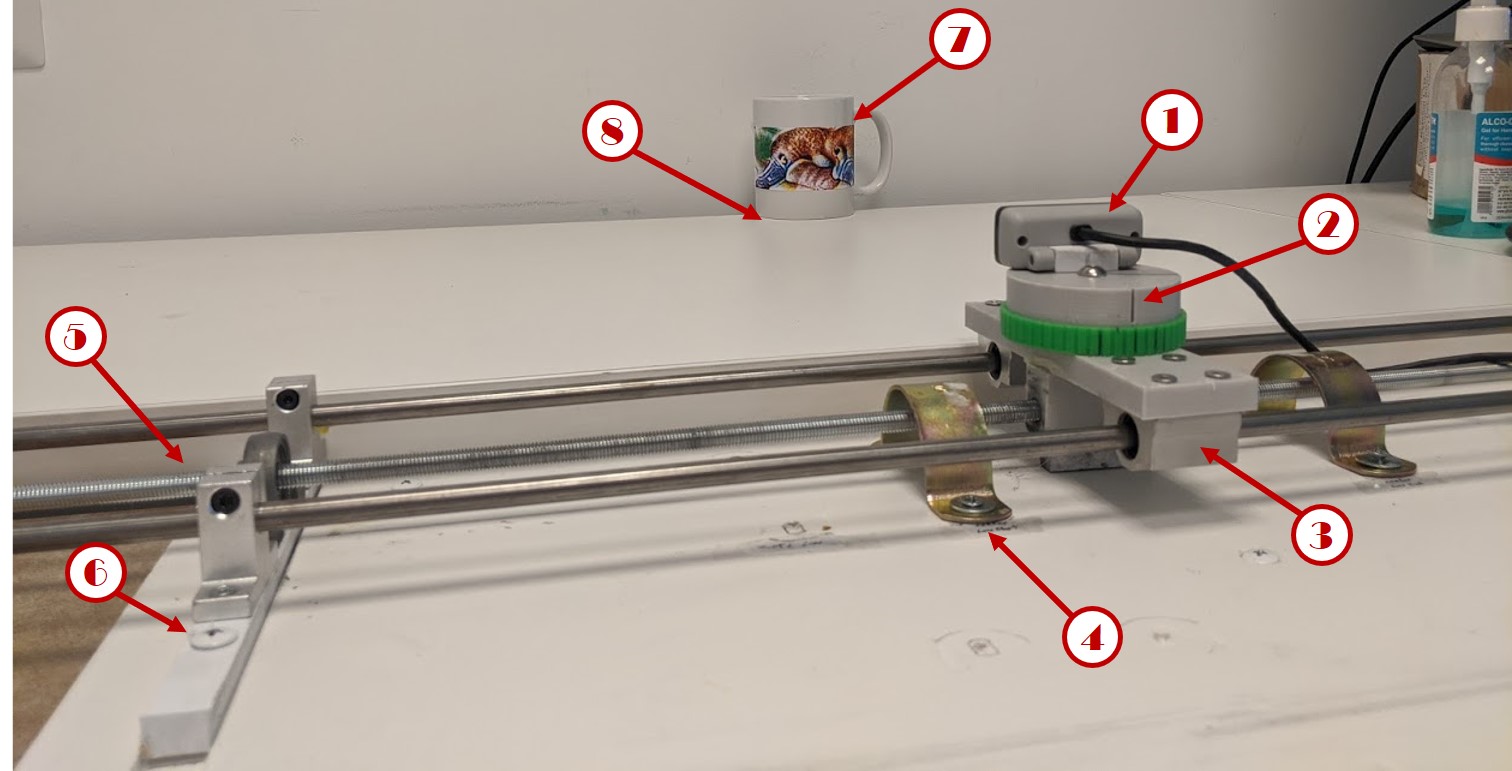}
    \caption{
    Our real-world evaluation setup:
    The position of the camera (1) is controlled by spinning a platform with gradations on its base (2).
    The platform is located on a cart (3), which moves within predefined ranges (4) on a spinning screw rod (5).
    The slider can be moved forward and backward into one of three predefined positions (6).
    We can rebuild the scene using the markings for the mug's location and position (7) and the patch's location on the mug (8).
    }
    \label{fig:rw-eval}
\end{figure*}

\begin{figure*}
    \centering
    \null\hfill
    \subfloat[Platform]{\label{fig:setup-closeup-platf}
        \includegraphics[width=0.32\linewidth]{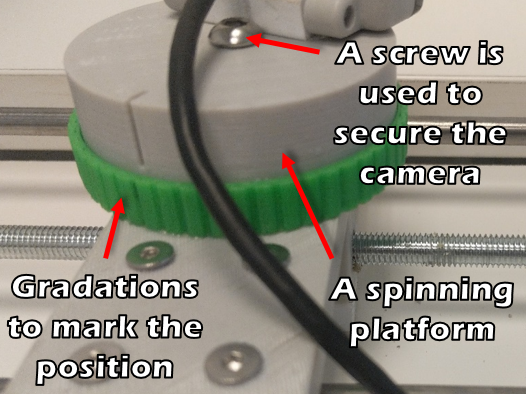}}\hfill
    \subfloat[Cart and ranges]{\label{fig:setup-closeup-cart}
        \includegraphics[width=0.32\linewidth]{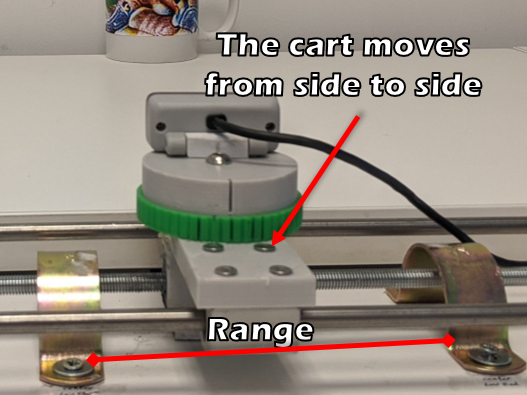}}\hfill
    \subfloat[Slider]{\label{fig:setup-closeup-slider}
        \includegraphics[width=0.32\linewidth]{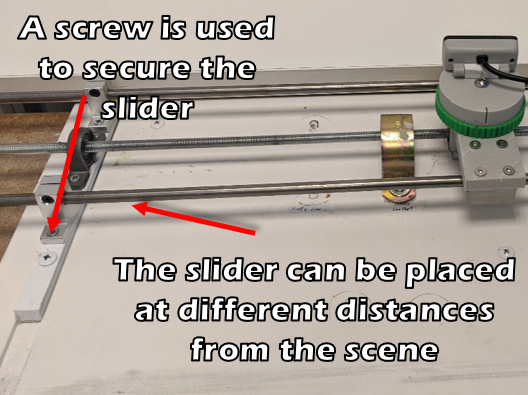}}
    \hfill\null
    
    \caption{
        A closeup of the three main components of the real-world evaluation setup:
        (a) the spinning platform for the camera, (b) the cart that moves the camera in front of the scene, and (c) the slider which can be placed at different distances from the scene.
    }
    \label{fig:setup-closeup}
\end{figure*}

To ensure that the real-world evaluation process is reproducible, we suggest using the evaluation setup that is presented in \fig{rw-eval}.
The setup includes a camera slider that can be placed at different distances from the scene and allows the camera to film the scene from predefined positions.
As shown in \fig{setup-closeup-platf}, the camera is placed on a spinning platform, which is located on top of a cart that moves the camera from side to side in front of the scene.
The position of the camera can be changed by spinning the platform to a specific angle, which is defined by marks on the gradations at the bottom of the platform.
Then, a screw is used to secure the camera in place to ensure that the camera's angle does not change during the experiment.
A motor spins a screw rod, which moves the cart across two metal rods at a constant speed.
We suggest defining small ranges on the slider, as shown in \fig{setup-closeup-cart}, from which the scene is filmed at different angles.
For example, after choosing a camera position, we identify the range on the slider in which the mug is visible to the camera; during the experiment, the cart moves only in the defined range to avoid filming irrelevant parts of the scene.
Each range should be defined based on the required observation area (from the center, left, or right sides of the slider), along with the corresponding position of the camera, and can be set by physically limiting the cart's movement or by configuring the motor's behavior.
Additionally, as shown in \fig{setup-closeup-slider}, the slider can be moved forward and backward across a grooved base to film the scene from different distances.
Finally, since the real-world scene might change, the location and position of each object, including the adversarial patch, must be marked, thus ensuring that the same scene can be re-evaluated. 
By physically marking each configuration, the same actions can be performed in future experiments and thus, achieve similar results.

\subsubsection{Our Evaluation Setup}
In this study, we use the the evaluation process described above, in both the digital space and real world, to evaluate the performance of the patches that we create using our framework.

\textbf{Digital space:}
We determine 3,360 different positions and locations for the camera around the object, based on the patch’s visibility in the digital replica; for each of them, we render an image and send it to the object classifier. 

\textbf{Real world:} 
We build the structure, as shown in \fig{rw-eval} and \fig{setup-closeup} and described above, with the following configuration:
The slider can be placed at one of three predefined distances from the scene (close, middle, and far).
We defined three ranges (left, center, and right) for each distance and mark each of the nine ranges using metal eye straps.
The location of the mug, the position of the mug's handle, and the place of the patch on the mug are marked, thus ensuring that the same scene is used throughout the various experiments.
Sketches and instructions for replicating our real-world evaluation setup are available in \citep{author_repository}.

\section{Results}
For each target class, we craft four patches: \textit{random, systematic, google,} and \textit{imagenet}.
The \textit{random} and \textit{systematic} patches are adversarial patches crafted by using our framework with the random or systematic sampling of the transformations during the rendering step.
The \textit{google} and \textit{imagenet} patches are made out of images of the target class that were obtained from a Google search or the ImageNet dataset respectively, and are used to ensure that the results are non-biased.
We note that the images used for the \textit{imagenet} patches were collected from the dataset that was used to train the classifier, Inception V3.
We also examine the classification results for a mug without a patch (\textit{clean}) and a patch with random pixel values (\textit{noise}).
For each target class and its four patches, we evaluate the percentage of images classified as the original class (Coffee Mug), target class, or other classes (out of the ImageNet dataset).

\subsection{Results for Evaluation in the Digital Space}
After changing the seed and initialization parameters, we craft \textit{random, systematic, google,} and \textit{imagenet} patches for the following classes: Armadillo, Hat, (bottle) Nipple, Platypus, Mask, Pencil Box, Syringe, Screw, Mousetrap, and Ladybug.
Then, we perform the evaluation process in the digital space for \textit{clean}, with the \textit{noise} patch, and each of the patches described.
All of the 3,360 images of the rendered replica for \textit{clean} and with the \textit{noise} patch are classified as the original class, Coffee Mug.
Table \ref{table:digital-results} summarizes the results for the remaining patches.

\begin{table*}[t]
\caption{
    The classification results (percentage) in the digital space for the original (Og), target (Tg), and other (Ot) classes.
    All of the images (100\%) of \textit{clean} and \textit{noise} are classified as the original class.
    }
\begin{center}\small
\begin{tabular}{|l|ccc|ccc|ccc|ccc|}\hline
    \multicolumn{1}{|l|}{} & \multicolumn{3}{c|}{Systematic} & \multicolumn{3}{c|}{Random} & \multicolumn{3}{c|}{Google} & \multicolumn{3}{c|}{Imagenet} \\ \hline
    \makecell{Target\\Class}         & Og       & Tg        & Ot      & Og      & Tg      & Ot     & Og       & Tg     & Ot     & Og        & Tg     & Ot      \\ \hline\hline
    Armadillo            & 0.3      & 99.5      & 0.2     & 0.8     & 98.5    & 0.7    & 98.1     & 0      & 1.9    & 98        & 0      & 2       \\ \hline
    Hat                  & 0.3      & 99.6      & 0.1     & 0.1     & 99.5    & 0.4    & 99.5     & 0      & 0.5    & 99.1      & 0      & 0.9     \\ \hline
    Nipple               & 0.1      & 97.6      & 2.3     & 0.4     & 93.1    & 6.5    & 99.6     & 0      & 0.4    & 100       & 0      & 0       \\ \hline
    Platypus             & 1        & 97.5      & 1.5     & 1.3     & 98      & 0.7    & 98.1     & 0      & 1.9    & 96.4      & 0      & 3.6     \\ \hline
    Mask                 & 1        & 99        & 0       & 6.9     & 92.9    & 0.2    & 99.7     & 0      & 0.3    & 99.7      & 0      & 0.3     \\ \hline
    Pencil Box           & 0        & 99.9      & 0.1     & 0.7     & 98.7    & 0.6    & 99.7     & 0      & 0.3    & 99.9      & 0      & 0.1     \\ \hline
    Syringe              & 0        & 98.9      & 1.1     & 0.8     & 95      & 4.2    & 99.2     & 0      & 0.8    & 97.6      & 0      & 2.4     \\ \hline
    Screw                & 0.1      & 99.8      & 0.1     & 0.6     & 97.6    & 1.8    & 100      & 0      & 0      & 97.9      & 0      & 2.1     \\ \hline
    Mousetrap            & 0        & 100       & 0       & 0.1     & 99.1    & 0      & 100      & 0      & 0      & 97.9      & 0      & 2.1     \\ \hline
    Ladybug              & 0        & 100       & 0       & 0       & 100     & 0      & 99.7     & 0      & 0.3    & 100       & 0      & 0       \\ \hline
\end{tabular}
\end{center}
    
    \label{table:digital-results}

\end{table*}

We expected that the \textit{imagenet} patches, which were taken from the dataset that was used to train Inception V3, would act like an adversarial patch, thus causing the scene to be classified as the target class.
Although the scene is simple, the \textit{google} and \textit{imagenet} patches, with clear images of the target class, do not affect the classifier. 
On average, 99\% of the rendered images of the replica with the \textit{google} and \textit{imagenet} patches are classified as the original class, and none of them are classified as the target class.
In contrast, more than 99\% and 97\% (on average) of the images are classified as the target class for the \textit{systematic} and \textit{random} patches respectively.
When comparing the two types of adversarial patches, the systematic sampling approach is significantly better at causing the scene to be classified as the target class than the random sampling approach (the p-value is 0.04 for a paired sample t-test).

\subsection{Results for Evaluation in the Real World}
For the real-world evaluation, we use the same patches we created in the digital space for the following target classes: Armadillo, Hat, Nipple, and Platypus.
Each of these patches, including the \textit{noise} patch, is printed on a piece of paper, cropped, and placed on the designated location on the coffee mug using transparent adhesive tape.
Then, we follow the setup presented in \fig{rw-eval} to take approximately 700 photos, from all nine ranges, of the real-world scene with each patch and \textit{clean}.
Similarly to the evaluation in the digital space, 100\% of the photos of the real-world scene for \textit{clean} and with the \textit{noise} patch are classified as the original class. 
Table \ref{table:rw-results} summarize the results for the remaining patches.

\begin{table*}[t]
\caption{
    The classification results (percentage) in the real world for the original (Og), target (Tg), and other (Ot) classes.
    All of the photos (100\%) of \textit{clean} and \textit{noise} are classified as the original class.
    }
\begin{center}\small
\begin{tabular}{|l|ccc|ccc|ccc|ccc|}\hline
    \multicolumn{1}{|l|}{} & \multicolumn{3}{c|}{Systematic} & \multicolumn{3}{c|}{Random} & \multicolumn{3}{c|}{Google} & \multicolumn{3}{c|}{Imagenet} \\ \hline
    \makecell{Target\\Class}         & Og       & Tg        & Ot      & Og      & Tg      & Ot     & Og       & Tg     & Ot     & Og        & Tg     & Ot      \\ \hline\hline
    Armadillo            & 1.5      & 92.8      & 5.7     & 2.1     & 95.3    & 2.6    & 95.8     & 0      & 4.2    & 99.7      & 0      & 0.3     \\ \hline
    Hat                  & 0        & 99.1      & 0.9     & 0.2     & 98.9    & 0.9    & 99.3     & 0      & 0.7    & 95.9      & 0      & 4.1     \\ \hline
    Nipple               & 0        & 98.7      & 1.3     & 0.7     & 92.7    & 6.6    & 100      & 0      & 0      & 99.1      & 0      & 0.9     \\ \hline
    Platypus             & 0.3      & 92.6      & 7.1     & 6.2     & 87      & 6.8    & 99.2     & 0      & 0.8    & 99.1      & 0      & 0.9     \\ \hline
\end{tabular}
\end{center}
    \label{table:rw-results}
\end{table*}

The results for the real-world evaluation are similar to the results obtained in the digital space.
For all target classes, the scene with the \textit{google} and \textit{imagenet} patches is mainly classified as the original class (for 98.5\% of the photos on average) and is never classified as the target class.
Similarly, the scene with the \textit{systematic} and \textit{random} adversarial patches is mainly classified as the target class; 
The average difference between the digital and real-world results is 5\%, and the actual difference never exceeds 7\%. 

We further examine the patches with the greatest difference between the digital space and real-world results.
The \textit{google} patch for the Armadillo target class and the \textit{imagenet} patch for Hat are classified as the original class in 98.1\% and 99.1\% of the images respectively, but only in 95.8\% and 95.9\% of the photos (respectively) of the real-world scene.
Both patches are classified as the Candle class from the ImageNet dataset (other) for photos taken from positions in which the mug's handle is less visible; therefore, the misclassification might stem from the cylindrical shape of the mug without the handle.
We also observe that the \textit{random} patch for the Platypus class is less effective when the camera is located far from the scene, in both the digital space and the real world.
Since the digital space evaluation reveals its weaknesses, the attacker can utilize this information to improve the patch by rendering more views in which the camera is located far from the scene.
This is an example of how the attacker can identify the patch's flaws in advance, learn how to improve the patch, and increase the chances of a successful attack.

\section{Resilience to Unexpected Transformations}
In realistic scenarios, the attacker cannot control the real-world scene, which means that by the time the attacker returns to the scene with the patch, the scene may have changed.
As we discussed, the attacker can use our framework to improve the patch's robustness to predictable changes, yet this is not the case for unexpected major transformations to the scene (e.g., the sudden removal of an object).
If the patch is only effective in the modeled scene and under expected transformations, then our framework is less feasible in real-world settings.
Therefore, we want to examine how unexpected transformations to the real-world scene affect the ability of adversarial patches created using our framework to fool the target neural network.
To do so, we use seven new transformations to change the real-world scene. 
The adversarial patches are not expected to be robust to those transformations, since they are not part of the digital replica.

\begin{figure*}
    \centering
    \null\hfill
    \subfloat[Different mugs]{\label{fig:transform-mugs}
        \includegraphics[width=0.40\linewidth]{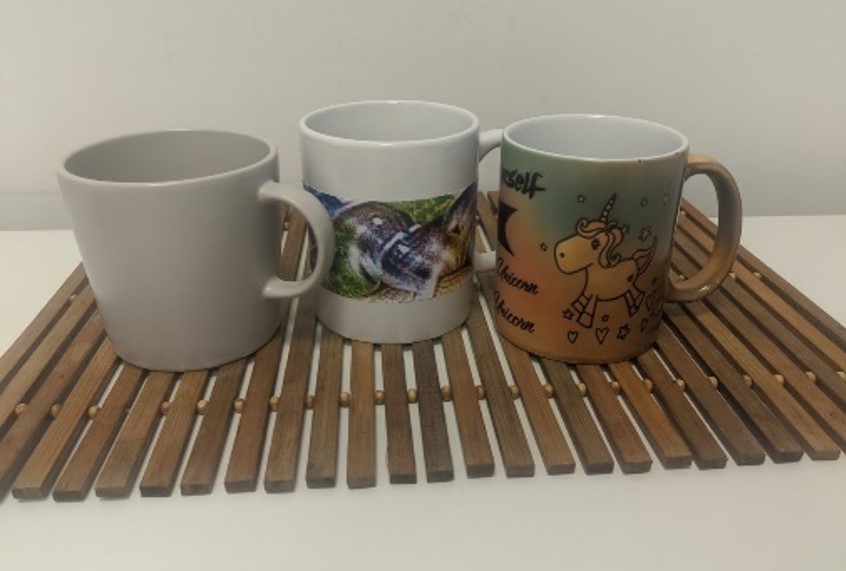}}\hfill
    \subfloat[Red mat]{\label{fig:transform-red}
        \includegraphics[width=0.40\linewidth]{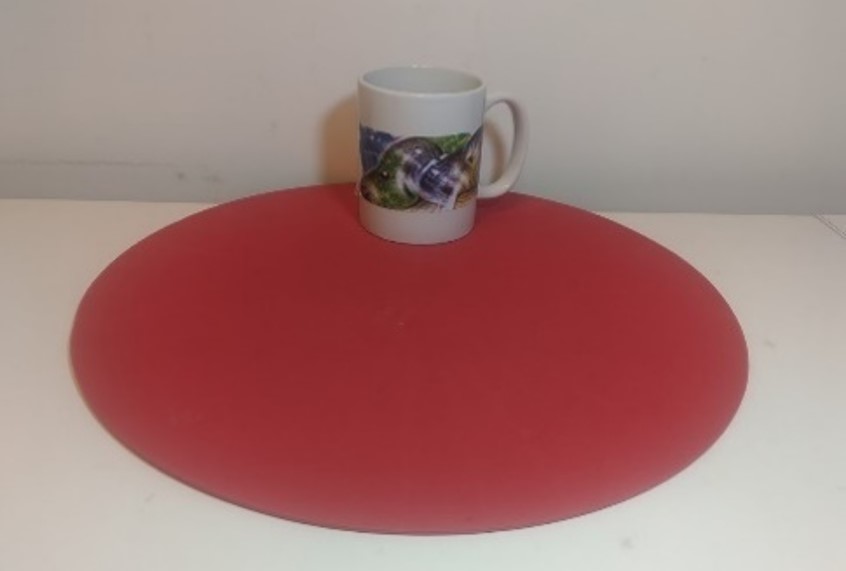}}
    \hfill\null
    
    \caption{
        Unexpected transformations to the real-world scene; (a) the three coffee mugs on the wooden mat used for \textit{wood} (from left to right: \textit{shape}, the original mug, and \textit{color}), and (b) \textit{red}: the original coffee mug with an adversarial patch on a red mat.
    }
    \label{fig:transform}
\end{figure*}

As shown in \fig{mugs-rw}, the original scene contains a white coffee mug with a handle on the right side and a patch placed in the middle of the mug, which is placed on an office desk.
We defined the seven unexpected transformations as \textit{up, down, red, wood, color, shape}, and \textit{flipped}.
\textit{Up} and \textit{down} are transformations to the patch's location on the mug: the patch is placed on the top and bottom of the mug, respectively.
We also change the surface on which the mug is placed to a red circle (\textit{red}) and a wooden mat (\textit{wood}).
For \textit{color} and \textit{shape}, we replace the original white mug:
for \textit{color}, we use a mug of the same shape but with a different color (a dark background with colorful illustrations), while for \textit{shape}, we use a mug of a similar color (light gray instead of white) which has a different shape. 
The \textit{shape} mug is shorter and cone-shaped, and has a smaller handle which is located at a higher position on the mug.
Finally, for \textit{flipped}, we rotate the mug by $180^\circ$.
In this experiment, for each transformation (some of the transformations are shown in \fig{transform}), we change the scene and perform the real-world evaluation process with the two adversarial patches (\textit{systematic} and \textit{random}) used for the Nipple target class. 
Table \ref{table:transformations} presents how each unexpected transformation affects the classification results of the real-world scene with the adversarial patches.

\begin{table*}[t]
\caption{
        The results of a real-world evaluation for the Nipple patches with seven unexpected transformations added to the scene. 
        The results indicate the percentage of photos classified as the original, target, or other classes.  
    }
\begin{center}
\begin{tabular}{|l||c|c|c||c|c|c|}\hline
    \multicolumn{1}{|c||}{} & \multicolumn{3}{c||}{Systematic} & \multicolumn{3}{c|}{Random} \\ \hline
    Transformation  & Original  & Target    & Other         & Original  & Target    & Other     \\ \hline\hline
    Up              & 0.6       & 88.9      & 10.5          & 2.2       & 77.9      & 19.9      \\ \hline
    Down            & 2.7       & 83.7      & 13.6          & 2.7	    & 87.8	    & 9.5       \\ \hline
    Red             & 1.9	    & 84.6	    & 13.5	        & 2.7	    & 76.5	    & 20.8      \\ \hline
    Wood            & 5.8	    & 68.9	    & 25.3	        & 2.4	    & 64.7	    & 32.9      \\ \hline
    Color           & 6.5	    & 39.1	    & 54.4	        & 5.5	    & 28.1	    & 66.4      \\ \hline
    Shape           & 0	        & 87.1	    & 12.9	        & 1.2	    & 71.7	    & 27.1      \\ \hline
    Flipped         & 2.7	    & 85.8	    & 11.5	        & 4.3	    & 55.1	    & 40.6      \\ \hline
    
\end{tabular}
\end{center}
    
    \label{table:transformations}
\end{table*}

The unexpected transformations cause the scene to be classified as the target class less often and more often as other classes. 
Additionally, the scene is classified as the original class in only 2.9\% of the photos (on average) and in no more than 6.5\% of the photos. 
This shows that the neural network fails to identify the mug even when there are major unexpected changes in the scene.
It seems that the patches are robust to changes in the patch's location on the mug; in \textit{up} and \textit{down}, the scene is classified as the original class for no more than 2.7\% of the photos.
However, changes in the object that the mug is placed on show mixed results; while the patches are more robust to \textit{red}, \textit{wood} is classified as the target class in less than 70\% of the photos, and in just up to 5.8\% of the photos as the original class.
We believe that the difference may stem from the shape of the mats: unlike the wooden mat, the red mat and the desk in the original scene have a solid shape.
Similarly, replacing the original mug also shows mixed results; both patches perform the worst on \textit{color} but perform well on \textit{shape}.
For \textit{color}, the scene is classified as the target class in less than 40\% of the photos and as the original class in up to 6.5\% of the photos.
We assume that the use of a mug with an inconsistent design (i.e., many colorful illustrations) reduces the performance of the patches.
Although the patches are the least robust to \textit{color}, the neural network still fails to identify the scene in more than 90\% of the photos.
In contrast, the scene with \textit{shape} is classified as the target scene in 71.7\% and 87.1\% of the photos for the \textit{random} and \textit{systematic} patches respectively, and the \textit{systematic} patch is never classified as the original class.
Additionally, we find that \textit{flipped} affects the adversarial patches differently: \textit{systematic} performs well, with 85.8\% of the photos classified as the target class, but for \textit{random}, only 55.1\% of the photos are classified as the target class.
Finally, in most cases, the \textit{systematic} patch performs better than the \textit{random} patch. 

This experiment was designed to examine whether using a digital replica limits the attack to a specific model of the scene, thus causing the adversarial patch to become ineffective when unexpected changes occur in the real world.
To do so, we performed noticeable, yet realistic, transformations to the original real-world scene: changes in an object's location and position, the replacement of an object, and adding a new object to the scene.
The results show that although the transformations reduce the attacker's ability to control the neural network's prediction, the patches can still fool the neural network so that it misclassifies the scene.
On average, the scene is not classified as the original label in more than 97\% of the photos, which supports the feasibility of our framework.
Therefore, the attacker can improve the attack's robustness to real-world transformations by creating a 3D replica of the scene (as we proposed), suffering just a minor reduction in the patch's fooling ability if unexpected changes occur in the target scene.

\section{Conclusions}
In this work, we demonstrated how 3D modeling techniques and tools can be used to craft inexpensive adversarial patches that are robust to real-world transformations. 
By creating a digital replica of the target scene, our method gives the attacker control of every aspect of the scene, including the objects, lighting, and more. 
The replica simulates the real-world scene, thus allowing the attacker to test and improve the attack without the risk of detection. 
We also demonstrate that such approach can improve the patch’s robustness to both expected and unexpected changes in the real-world scene. 
Additionally, we present an evaluation process that enables other researchers to reproduce our experiments and validate our results.
We believe that such an evaluation process can be used in future studies and contribute to replicating, examining, and improving other real-world attacks. 

In future work, we plan to improve the suggested framework by adding 3D elements (e.g., reflection and normal maps) to support more complex scenes.
Then, we will use it to tailor an attack for other domains, like the facial recognition domain, where our framework can improve individuals' privacy concerning such systems.
We also aim to create imperceptible perturbations that attract less attention. 
Finally, our results suggest that using the EOT framework with systematic sampling might be better than random sampling, and we plan to perform additional experiments to examine this further.

\section*{Acknowledgment}
We gratefully acknowledge Matan Yesharim for his major contribution to the development of both the attack and evaluation process. 
We also thank Boris Zadov for his professional expertise. 
Finally, we extend a special thank you to Mathov Designs for designing and building the real-world evaluation setup and for allowing us to publish their designs which will assist the research community worldwide.

\bibliographystyle{elsarticle-num-names}
\bibliography{cas-refs}





\end{document}